\documentclass[runningheads]{llncs}
\usepackage{graphicx}

\usepackage{multirow}
\usepackage{booktabs}
\usepackage{caption}
\usepackage[caption=false]{subfig}

\usepackage[bottom]{footmisc}

\begin{document}
\title{Revisiting the Shape-Bias of Deep Learning\\for Dermoscopic Skin Lesion Classification}
\titlerunning{Revisiting the Shape-Bias in Dermoscopy}

\author{Adriano Lucieri\inst{1,2}\orcidID{0000-0003-1473-4745} \and
Fabian Schmeisser\inst{1}\orcidID{0000-0001-8222-7900} \and
Christoph Peter Balada\inst{2}\orcidID{0000-0003-0307-7866} \and
Shoaib Ahmed Siddiqui\inst{2}\orcidID{0000-0003-4600-7331} \and
Andreas Dengel\inst{1,2}\orcidID{0000-0002-6100-8255} \and
Sheraz Ahmed\inst{2}\orcidID{0000-0002-4239-6520}}
\authorrunning{A. Lucieri et al.}

\institute{Department of Computer Science,\\Technische Universit\"at Kaiserslautern, \\Erwin-Schrödinger-Straße 52, 67663 Kaiserslautern, Germany \and
Smart Data and Knowledge Services (SDS),\\
German Research Center for Artificial Intelligence GmbH (DFKI), \\Trippstadter Straße 122, 
67663 Kaiserslautern, Germany\\
\email{firstname.lastname@dfki.de}\\
}
\maketitle

\begin{abstract}
It is generally believed that the human visual system is biased towards the recognition of shapes rather than textures.
This assumption has led to a growing body of work aiming to align deep models' decision-making processes with the fundamental properties of human vision.
The reliance on shape features is primarily expected to improve the robustness of these models under covariate shift.
In this paper, we revisit the significance of \textit{shape-biases} for the classification of skin lesion images.
Our analysis shows that different skin lesion datasets exhibit varying biases towards individual image features.
Interestingly, despite deep feature extractors being inclined towards learning entangled features for skin lesion classification, individual features can still be decoded from this entangled representation. 
This indicates that these features are still represented in the learnt embedding spaces of the models, but not used for classification.
In addition, the spectral analysis of different datasets shows that in contrast to common visual recognition, dermoscopic skin lesion classification, by nature, is reliant on complex feature combinations beyond \textit{shape-bias}.
As a natural consequence, shifting away from the prevalent desire of shape-biasing models can even improve skin lesion classifiers in some cases.

\end{abstract}

\section{Introduction}
\label{sec:introduction}

For over a decade now, Deep Neural Networks (DNNs) outperformed conventional techniques in various research areas including language translation~\cite{sutskever2014sequence}, image classification~\cite{krizhevsky2012imagenet} and image synthesis~\cite{goodfellow2014generative}.
Although new state-of-the-art performances are being reported continuously for areas like Melanoma detection~\cite{tang2022fusionm4net,hasan2022dermoexpert}, an unconstrained application of Deep Learning (DL) in real-world, high-stakes medical decision-making is still considered questionable due to a lack of robustness and intelligibility.
Several works have revealed weak spots of the current technology like the presence of adversarial examples~\cite{szegedy2014intriguing}, the influence of distribution shifts~\cite{quinonero2008dataset} and the bias-variance tradeoff~\cite{geman1992neural}.

Geirhos et al.~\cite{geirhos2018imagenet} disproved the widespread \textit{shape hypothesis}, which states that Convolutional Neural Networks (CNNs) hierarchically combine lower-lever features into higher-level features for generating the final predictions.
Instead the authors propose the \textit{texture hypothesis}, stating that an inherent \textit{texture-bias} in the dataset can lead to a lack of robustness in CNNs.
Similarly, other works~\cite{yin2019fourier} have reported a higher importance of texture-like high-frequency input features, which aligns with the vulnerability to high-frequency adversarial attacks~\cite{wang2020towards}.
Through expensive modification of the training dataset, exchanging the dataset's \textit{texture-bias} to a \textit{shape-bias}, the authors of~\cite{geirhos2018imagenet} achieve improved classification robustness.
Along similar lines, recent works~\cite{chen2021amplitude,xu2021fourier} exploit the idea that the phase spectrum of a Fourier-transformed image mainly encodes semantic information resembling edges and outlines used by humans for object identification.
The authors propose different data augmentation strategies for improved robustness, inducing explicit focus on the phase spectra of images, shifting the networks' focus towards shape information.
The common idea in these works is the explicit alignment of a network's non-functional requirements with those used in analytical, human decision-making (i.e. focusing on shape more than on texture).

Despite recent efforts towards suppressing high-frequency texture features in DL, Ilyas et al.~\cite{ilyas2019adversarial} argue that datasets can contain robust features which are indeed imperceptible to humans.
This alternative perspective is particularly interesting when dealing with complex medical problems which are yet to be fully understood by human experts and cannot be easily solved through intuition.
One of such high-stakes use-cases of DL in medicine is the classification of Melanoma, which is mainly driven by non-analytic clinical reasoning (i.e. pattern analysis~\cite{kittler2016dermatoscopy}).

The statistical relevance of shapes, textures and colors in dermoscopic images for Melanoma detection has been extensively investigated in different studies.
Marques et al.~\cite{marques2012role} reported that color and texture features individually have a high relevance for skin lesion classification, but their combination is even more informative.
In other works~\cite{ruela2013color,barata2014bag} the superior role of color features is reported.
Ruela et al.~\cite{ruela2013shape} investigate the importance of shape features, concluding that, although shape is relevant for classification, the use of texture and color descriptors is more effective.
Beyond texture, shape, and color, other studies indicate a high relevance of spectral features for predictive performance~\cite{betta2006dermoscopic,lopez2021multi}.
However, the influence of individual features, as well as the effect of the \textit{shape-bias} on DL-based skin lesion classifiers has not yet been explored.

In this paper, we revisit the \textit{shape-bias} and it's effect on the analysis of dermoscopic images using DL-based models.
To that end, we explore the relevance of individual image features known to be relevant in dermoscopy (i.e. \textit{Texture}, \textit{Shape}, and \textit{Color}).
A spectral analysis on different datasets is performed to investigate the distribution of relevant image features in the spectral domain, and to revisit the effectiveness of robustness methods enforcing explicit \textit{shape-bias} on deep feature extractors.
Lastly, a new variant of the Amplitude-Phase Recombination~\cite{chen2021amplitude} robustness method is introduced, which is more aligned with the complex needs of dermoscopic skin lesion analysis.
We argue that the current trend of focusing network robustness in Deep Learning purely on the \textit{shape-bias} is to narrow-minded, and that medical imaging tasks (like dermoscopy) in particular, have radically different requirements when it comes to non-functional properties of their decision-making.

Section~\ref{sec:background} gives a brief introduction into the notions of \textit{Texture}, \textit{Shape}, and \textit{Color} in dermoscopy and describes image ablations used to isolate these different features.
The datasets used in throughout our work, as well as the general experimentation setting is outlined in section~\ref{sec:datasets_and_methodology}.
In section~\ref{sec:isolated_feature_relevance}, the individual importance of isolated image features is investigated, and their encoding in the DL-based models' feature space.
A spectral analysis on different datasets is performed in section~\ref{sec:spectral_analysis}, followed by an investigation of shape-focused robustness methods.
Finally, the results are discussed in section~\ref{sec:discussion}, followed by the concluding remarks.

\section{Definition and Isolation of Texture, Shape, and Color in Skin Lesions}
\label{sec:background}

To properly investigate the influence of individual image features, we need to isolate image features and feature combinations from the input images. 
We follow the previous lines of work and concentrate on the \textit{Texture}, \textit{Shape} and \textit{Color} features as the main components descriptive of skin lesion images.
First, the individual features are briefly defined, based on the relevant literature. 
Then, the transformations achieving the different feature isolations are elaborated and presented.

\paragraph{\textbf{Texture}} Marques et al.~\cite{marques2012role} define textures in skin lesions as conveying "\textit{information about the differential structures (pigment network, dots, streaks, etc) present in the lesion}".
We therefore argue that textures are solely encoded in structures such as fine edges, and color contrasts.

\paragraph{\textbf{Shape}} Shape descriptors are computed in~\cite{ruela2013shape} based on the segmented lesion outline.
The measures include simple shape descriptors such as the lesion's area, compactness, and rectangularity, but also more advanced features such as symmetry-related features and moment invariants.
In this work, we define the shape of a lesion by the size and area of a lesion's segmentation, as well as the regularity and overall shape of it's outline. 
For the sake of simplicity, we omit information regarding the smoothness of a lesion's transition.

\paragraph{\textbf{Color}} In~\cite{marques2012role}, color features are defined as containing "\textit{information about the color distribution and number of colors in the skin lesion}".
Ruela et al.~\cite{ruela2013color} compute different color descriptors based on different color spaces.
Most descriptors contain only information about the quantitative distribution of color in an image, whereas one descriptor also encodes information about the spatial location.
We argue that color information is not only encoded in the absolute color values, but also in the contrast information of broader surfaces.
However, we do not regard spatial color information for the sake of simplicity.

\subsection{Feature Ablations}

\begin{figure}[t!]
\includegraphics[width=\textwidth]{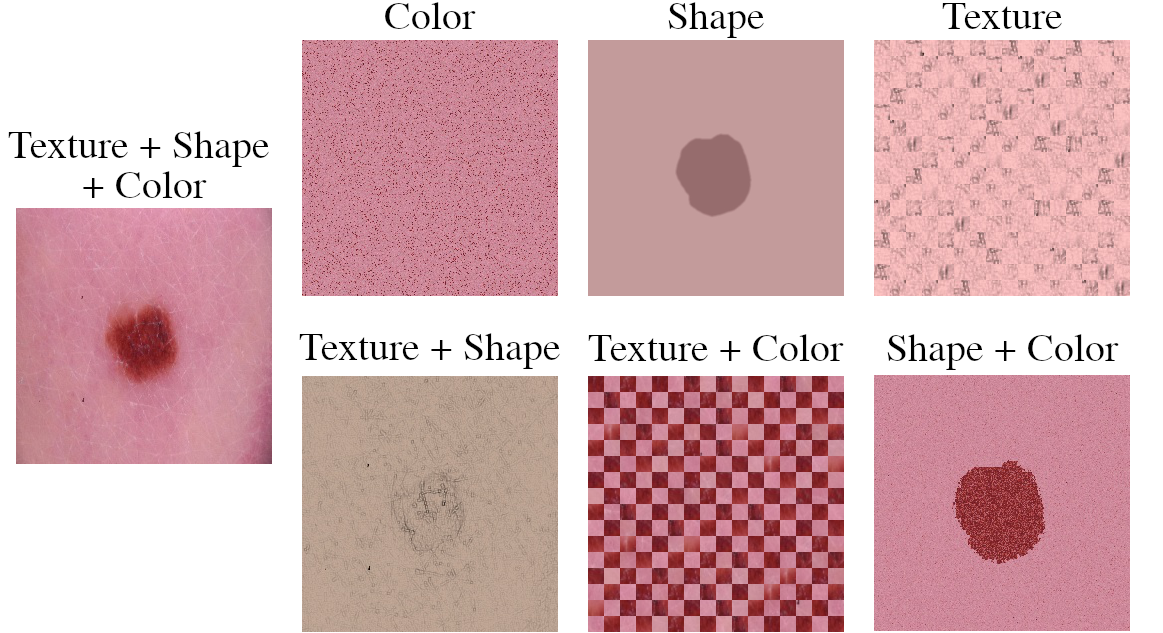}
\caption{Illustration of the different augmentations designed to isolate individual image features in skin lesion images. The original image is provided on the left. The first row shows augmentations for isolated images, while the second row shows combinations of two individual features.} 
\label{fig:ablations}
\end{figure}

To isolate the effect of individual features, we design different data augmentation strategies, each representing one of the seven unique feature combinations.
An illustration of the different transformations is provided in figure~\ref{fig:ablations}.
% Original
The original images combine all three features and serve as the baseline.
% Color
An ablation representing only the \textit{Color} feature is obtained by randomly scrambling the spatial ordering of individual image pixels, while preserving the sample's original color distribution.
% Texture + Shape
A combination of \textit{Texture} and \textit{Shape} is achieved by explicitly removing the \textit{Color} cues. 
This is done by changing the style of a sample to a sketch-like image, in order to remove color value and contrast information, while maintaining the characteristic edges necessary to identify textures and shapes.
% Shape
A DL-based segmentation model\footnote{BA-Transfomer architecture proposed by Wang et al.~\cite{wang2021boundary}, trained on ISIC2016-2018 challenge datasets.} is tuned for the computation of lesion segmentation maps, representing the isolated \textit{Shape} feature.
% Texture + Color
The \textit{Shape} feature is removed from the original image with the help of the segmentations as well.
Therefore, the segmentation map is divided in equally-sized patches to identify image regions containing information about the lesion's outline (i.e. containing both lesion and background pixels).
After removing all outline patches, a new image is assembled by alternatingly sampling random patches from the lesion and the background, to ensure that no information about the lesion's absolute size is retained.
A similar procedure is followed to obtain the isolation of the \textit{Texture} feature.
Instead of scrambling image patches of the original image, the sketch transformation is used in order to spare color information.
A combination of \textit{Shape} and \textit{Color} is obtained by separately scrambling the spatial ordering of individual image pixels within the lesion region, and the background.
Images with radical domain shifts (through sketch transformation or segmentation) are shaded with the channel-wise average of the dataset's color.

\section{Datasets and Methodology}
\label{sec:datasets_and_methodology}

\subsection{Datasets}

\paragraph{\textbf{ISIC \& ISIC-b}} The International Skin Imaging Collaboration (ISIC) organized several skin cancer classification challenges over the last decade.
The challenge datasets are hosted on the ISIC's online archive\footnote{https://www.isic-archive.com/}, which is the largest public database of dermoscopic skin images to date.
The complete archive consists of over 69.000 clinical and dermoscopic images of different provenance.
We follow Cassidy et al.~\cite{cassidy2022analysis}, who propose a duplicate removal strategy for the ISIC challenge datasets to avoid overlap between training and evaluation sets.
For experimentation, we combine all duplicate-free challenge training sets and generate new training, validation and testing splits under stratification.

The complete ISIC dataset has annotations for eight classes, i.e. Actinic Keratosis (AK), Basal Cell Carcinoma (BCC), Benign Keratosis (BKL), Dermatofibroma (DF), Melanoma (MEL), Nevus (NV), Squamous Cell Carcinoma (SCC) and Vascular Lesions (VASC).
We generate a multi-class variant of the dataset (henceforth referred to as \textit{ISIC}) comprising of 23.868 training, 2.653 validation and 2.947 testing samples.
In addition, a binary variant consisting of only NV and MEL samples (henceforth referred to as \textit{ISIC-b}) is generated comprising of 10.543 training, 4.519 validation and 6.456 testing samples.

\paragraph{\textbf{D7P \& D7P-b}} The seven-point checklist criteria dataset (\textit{D7P}) proposed in~\cite{kawahara2018seven} consists of clinical and dermoscopic images of 1.011 skin lesions.
Each image is annotated with regards to its diagnostic class, several dermoscopic criteria as well as further clinical data.
In this work, we only consider the subset of dermoscopic images along with the respective annotations of dermoscopic criteria and final diagnosis.
We follow the original work, categorizing the fine-grained annotations into BCC, NV, MEL, Sebbhoreic Keratosis (SK) as well as a miscellaneous (MISC) classes.
Again, we generate a stratified binary variant of only NV and MEL samples (henceforth referred to as \textit{D7P-b}) comprising of 371 training, 183 validation and 273 testing samples. For concept detection experiments, we follow the same splitting procedure for each dermoscopic concept.

\subsection{Experimental Setup}
If not mentioned otherwise, all experiments are conducted with a ResNet50, pre-trained on \textit{ImageNet}.
Training is conducted using softmax cross-entropy loss and AdamW optimizer.
The learning rate and weight decay are determined by hyperparameter tuning on the baseline setting.
A plateau learning rate scheduler is used in conjunction with an early stopping scheme to ensure convergence of the models.
Each training and respective evaluation is run $10$ times with varying random seeds to ensure significance of the reported results.\footnote{Reproducible code available on GitHub https://github.com/adriano-lucieri/shape-bias-in-dermoscopy}

\section{Deep Feature Extractors for Dermoscopy Can Encode Disentangled Features}
\label{sec:isolated_feature_relevance}

In this section, we perform an extensive study on the influence of individual features for the DL-based classification of dermoscopic skin lesions.
We show that even within the dermoscopy domain, biases are dataset-dependent.
Moreover, we show that although feature extractors are inclined towards learning entangled features, last-layer retraining can recover at least some features successfully.

\subsection{Different Dermoscopic Skin Lesion Datasets have Different Biases}
\label{sec:tsc_datasetwise}

In a first experiment, we train separate classifiers on the individual ablations introduced in section~\ref{sec:background}.
By training and testing on a specific ablation, we want to quantify the CNN-based feature extractor's capability to leverage individual, interpretable features from the images.
We consider both binary and multiclass skin lesion datasets to account for possible effects of varying task complexity. 

\begin{figure}[t!]
\includegraphics[width=\textwidth]{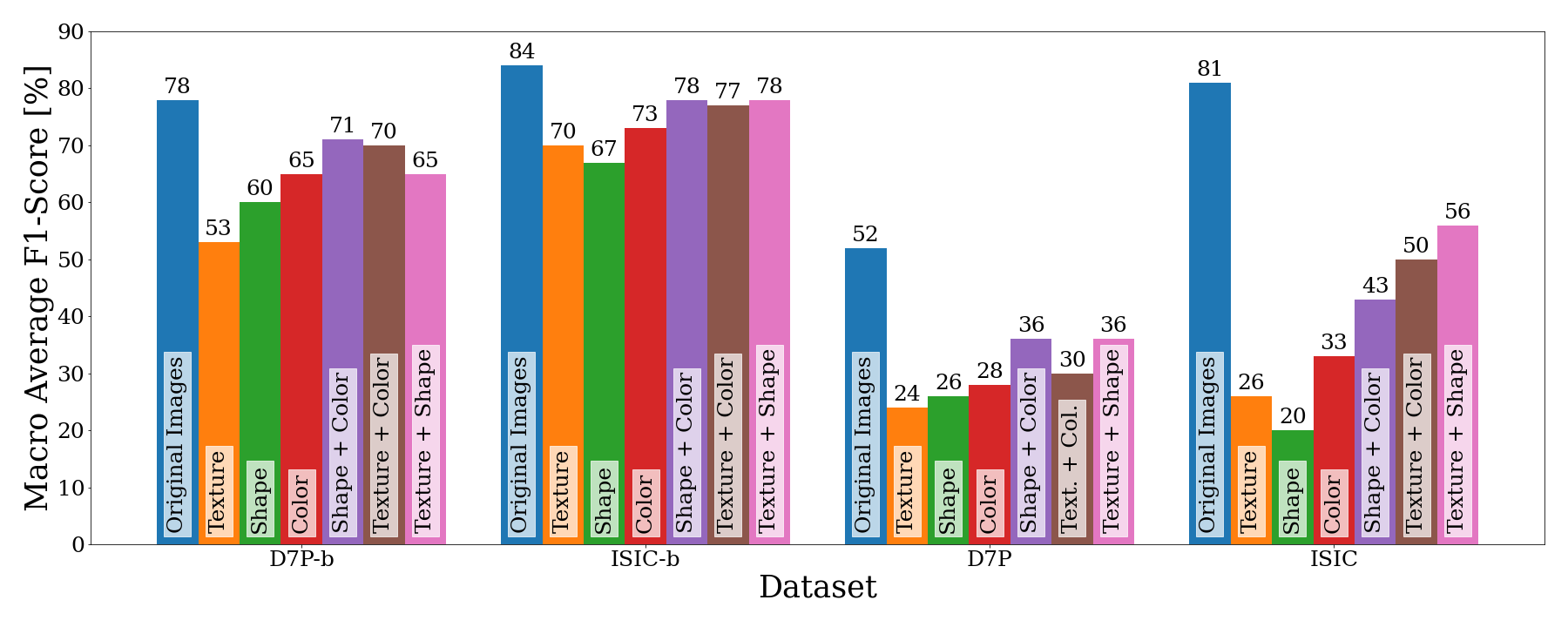}
\caption{Macro averaged F1-scores from models trained and tested on individual \textit{Texture}, \textit{Shape} and \textit{Color} feature ablations. The first bar of each group represents the reference F1-score achieved by models trained on unaltered data, followed by the three feature isolation and feature removal ablations.} 
\label{fig:tsc}
\end{figure}

Figure~\ref{fig:tsc} shows the macro averaged F1-scores on the test set of the respective ablations.
When providing only single features in isolation, we observe that both binary and multiclass datasets based on \textit{D7P}, show a stronger bias towards \textit{Shape}, whereas the \textit{ISIC} datasets are more sensitive with respect to \textit{Texture}.
This is indicated by the lower decrease in F1-scores when training on the respective isolated feature.
It can also be observed that \textit{Color} is the most important of all three features, resulting in the lowest performance decline.

Training on a pair of two individual features can be considered as the removal of the absent feature, and therefore serves as an inverted indicator for feature importance.
The previous observations are also confirmed by the results of removing \textit{Texture} and \textit{Shape}, except for \textit{ISIC-b} which is influenced almost equally.
Surprisingly, removing the \textit{Color} did not indicate similar relevance as indicated by the isolation experiment.
A reason for this behaviour could be that the combination of \textit{Shape} and \textit{Texture} information forms stronger higher-level features as compared to combinations including \textit{Color}.

\subsection{Dermoscopic Skin Lesion Classifiers Learn Entangled Features}

\begin{figure}[t!]
\includegraphics[width=\textwidth]{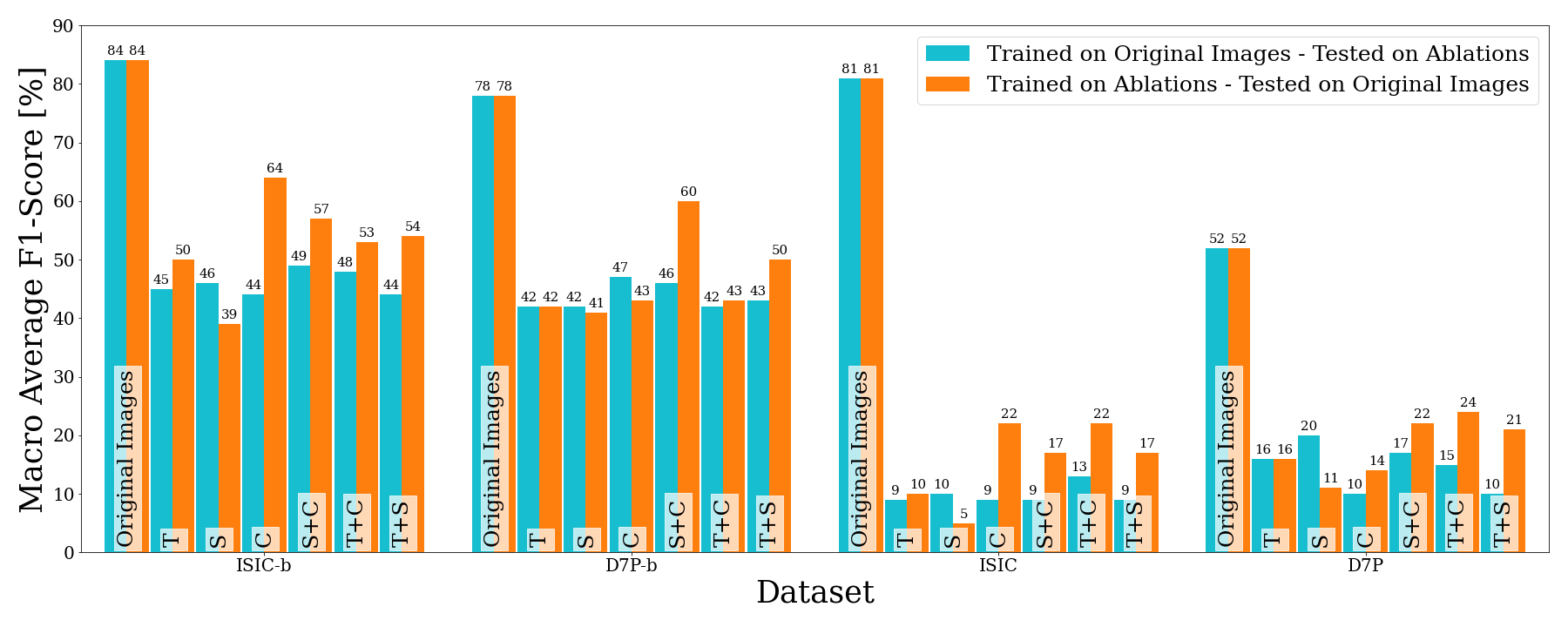}
\caption{Comparison of macro averaged F1-scores from models trained and tested on different data ablations. Blue bars show results of models trained on original images, and tested on different ablations. Orange bars show results of models trained on individual ablations, and tested on original images. \textit{T}, \textit{S}, \textit{C} refer to \textit{Texture}, \textit{Shape} and \textit{Color} features, respectively.} 
\label{fig:feature_transfer}
\end{figure}

The previous results showed that for some datasets, decent classification performances can be achieved even if only one or two features are present in the data.
We now investigate the ability of trained classifiers to transfer their features in ablated scenarios.
Therefore, the performance of each classifier trained in the previous experiment is measured across all ablations, as well as the original data.

Figure~\ref{fig:feature_transfer} shows a comparison of the results obtained when evaluating the models trained on the original input data on the test sets of all individual ablations (blue bars), with the inverse case, where different models trained on individual ablations are evaluated only on the original test set (orange bars).
The data clearly shows that the baseline classifier is unable to properly transfer it's learned features to the classification of ablated data, representing isolated input features.
However, the increased scores of the ablation-trained classifiers on the original data indicates the validity of the features even on unablated data.
Hence, we conjecture that the baseline models are not capable of making decisions purely based on the remaining features, but are instead overrelying on an entangled representation of different features.

\subsection{Feature Extractors are only partially Feature Biased}
\label{sec:dfr}

\begin{figure}[t!]
\includegraphics[width=\textwidth]{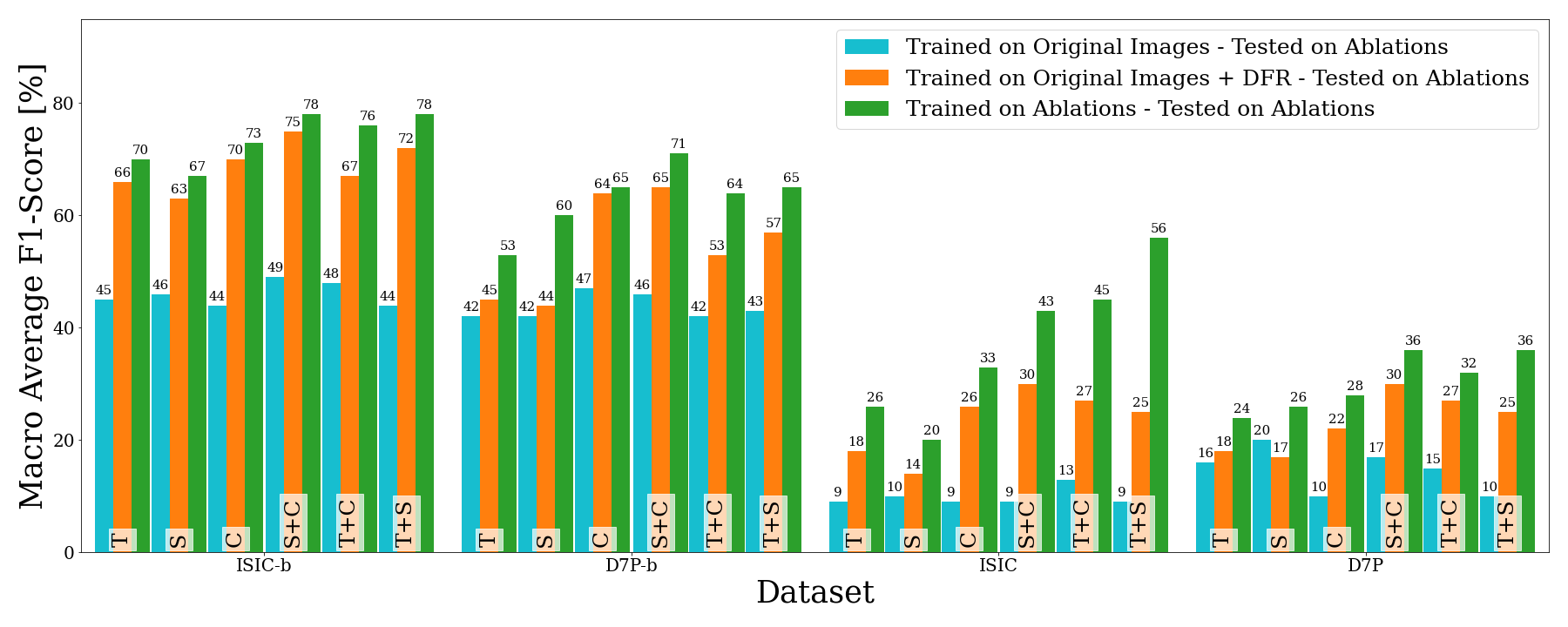}
\caption{Comparison of macro averaged F1-scores from models tested on different feature ablations. Blue bars show results of a baseline model trained on original images, and tested on all ablations. Orange bars show results of the same models after \textit{Deep Feature Reweighting}. Green bars show results of models trained and tested on ablations. \textit{T}, \textit{S}, \textit{C} refer to \textit{Texture}, \textit{Shape} and \textit{Color} features, respectively.} 
\label{fig:tsc-DFR}
\end{figure}

Inspired by \textit{Deep Feature Reweighting} (DFR) proposed in~\cite{kirichenko2022last}, we explore the level of entanglement in the skin lesion classifiers' feature spaces.
We again utilise the models trained on the original images from section~\ref{sec:tsc_datasetwise} but retrain the fully connected classification layers using the respective feature-ablated training and test sets.
In figure~\ref{fig:tsc-DFR}, the macro averaged F1-scores of baseline models with naive transfer, and DFR is compared to the accuracies of the models trained end-to-end on the ablated data.

It can be seen that in some cases (e.g. \textit{Color}, \textit{Shape + Color} and partially \textit{Texture + Color} as well as \textit{Texture + Shape} for \textit{ISIC-b}) DFR is able to achieve classification performances near the ideal values represented by the results of models trained and tested on ablations. 
For models trained on different skin lesion datasets, the successfully recovered features vary significantly. 
In contrast to \textit{D7P} and \textit{D7P-b} models, where only \textit{Color} could be recovered, models trained on \textit{ISIC-b} were able to recover all individual features to a sufficient degree.
A similar observation can be made when inspecting the results of multiclass \textit{ISIC} trained models.
Another striking observation is that combinations of two features resulted in higher DFR performance across all datasets. The results indicate that the feature extractor is always inclined towards learning entangled features.
However, the fact that individual features are additionally encoded, particularly in more complex and feature-rich datasets, confirms the results reported in~\cite{kirichenko2022last}.
This also suggests that an abundance of mostly redundant features in a dataset allows networks to learn alternative, isolated representations.

\section{Dermoscopy Relies on Complex Feature Combinations in Spectral Domain}
\label{sec:spectral_analysis}
In this section, we investigate the difference between feature entanglement in skin lesion classification tasks and common visual recognition. 
For comparability, we consider the distinction of features in the spectral domain, which has been commonly examined in previous studies~\cite{yin2019fourier,chen2021amplitude,xu2021fourier}.
We show that, compared to conventional visual recognition tasks, dermoscopy is more reliant on features across both amplitude and phase spectra.
To contrast the spread of dermoscopic features to those used in common visual recognition tasks, we utilize two subsets of \textit{ImageNet}~\cite{deng2009imagenet}, namely \textit{Imagenette} and \textit{Imagewoof}\footnote{https://github.com/fastai/imagenette}.

\subsection{Dermoscopy Features are Spread over Phase and Amplitude}
\label{sec:phase-amplitude-dependence}
First, we implement phase- and amplitude-randomization augmentations which are applied to train, validation and test images alike.
Phase-randomization (i.e. \textit{Amplitude-Only}) is applied by replacing the phase spectrum of an image, with the phase spectrum of a randomly sampled image of Gaussian noise in Fourier domain.
The same procedure is followed analogously for Amplitude-randomization (i.e. \textit{Phase-Only}).

\begin{table}[t!]
\caption{Resulting testing accuracies from retraining ResNet50 with different spectral randomizations. 
For both spectral randomizations, the performance decrease with respect to the baseline training is provided in a separate row.
}
\label{tab:randomization}
\begin{tabular}{p{0.22\textwidth}p{0.1\textwidth}p{0.1\textwidth}p{0.1\textwidth}p{0.1\textwidth}p{0.16\textwidth}p{0.16\textwidth}}
\toprule
\textbf{Dataset}            & \textbf{D7P-b} & \textbf{D7P} & \textbf{ISIC-b} & \textbf{ISIC}   & \textbf{Imagenette} & \textbf{Imagewoof} \\ \midrule
Baseline & \phantom{-}82.01 & \phantom{-}69.13 & \phantom{-}88.45 & \phantom{-}85.93 & \phantom{-}97.77 & \phantom{-}92.53 \\ \cline{2-7}
Amplitude-Only & \phantom{-}69.71 & \phantom{-}36.53 & \phantom{-}78.42 & \phantom{-}52.94  & \phantom{-}50.28 & \phantom{-}26.86 \\
 \phantom{----}$\Delta$ Baseline & -15.00 & -47.16 & -11.34 & -38.39 & -48.57 & -70.97 \\ \cline{2-7}
Phase-Only & \phantom{-}67.66 & \phantom{-}45.45 & \phantom{-}80.78 & \phantom{-}69.29 & \phantom{-}91.98 & \phantom{-}81.53 \\
 \phantom{----}$\Delta$ Baseline & -17.50 & -34.25 & -08.67 & -19.36 & -05.92 & -11.89 \\ \bottomrule
\end{tabular}
\end{table}

Table~\ref{tab:randomization} shows the test results of models trained in the \textit{Amplitude-} and \textit{Phase-Only} settings in comparison to the baseline classification accuracies over five different datasets.
It can be observed that \textit{Amplitude-Only} always leads to comparatively high deterioration of accuracy for all datasets. 
However, \textit{Phase-Only} results in comparable accuracy drops over all skin lesion datasets, even causing a higher relative performance decrease for \textit{D7P-b}.
Both visual recognition datasets instead show a significantly higher decrease in accuracy when providing only amplitude information, as compared to the \textit{Phase-Only} setting.

The results indicate that skin lesion datasets rely heavily on both amplitude- and phase-spectra, therefore potentially considering a more complex composition and variety of features beyond simple shape information.
In contrast, both \textit{ImageNet} subsets show a significant bias towards the phase spectra of the images. 
The higher drop in accuracy when training \textit{Imagewoof} \textit{Phase-Only} indicates that a combination of phase and color is extremely important to achieve high performance in some classes, although phase being mostly sufficient.
Interestingly, an inspection of the confusion matrices reveals that in the \textit{Phase-Only} setting, networks tend to most often confuse \textit{Beagles} with \textit{English Foxhounds}, which share many features in their physique.
A similar observation was made for the \textit{Amplitude-Only} setting, where \textit{Rhodesian Ridgebacks} and \textit{Dingos} where confused most often, sharing a similar fur color.

\subsection{Focusing on Amplitude can Improve Performance}
Amplitude-Phase Recombination (\textit{APR}) has been proposed in~\cite{chen2021amplitude} as a method to increase the robustness of DL classifiers by focusing the feature extraction on the phase-spectrum of images. 
$\mathcal{F}_x$ is the spatial Discrete Fourier Transform (DFT) of an image $x$ over each individual channel.
This frequency representation can be decomposed into an amplitude component ($\mathcal{A}_x$) and a phase component ($\mathcal{P}_x$) as follows:
\begin{equation}\small
	\setlength{\abovedisplayskip}{3pt}
	\setlength{\belowdisplayskip}{3pt}
	\mathcal{F}_x = \mathcal{A}_x \otimes e^{i \cdot \mathcal{P}_x},
	\label{eqn:freq}
\end{equation}
Amplitude-Phase Recombination for pair samples (\textit{APR-P})~\cite{chen2021amplitude} augments a given input image $x_j$ by replacing its amplitude spectrum $\mathcal{A}_{x_j}$ with the spectrum of another randomly selected image from the batch ($\mathcal{A}_{x_k}$).
\begin{equation}\small
	\setlength{\abovedisplayskip}{3pt}
	\setlength{\belowdisplayskip}{3pt}
	x_{j,aug} = iDFT(\mathcal{A}_{x_k} \otimes e^{i \cdot \mathcal{P}_{x_j}}),
	\label{eqn:freq}
\end{equation}
Instead, we propose two new variations of \textit{APR-P}, namely Amplitude-Focused \textit{APR-P} (\textit{AF-APR-P}) and Mixed \textit{APR-P} (\textit{Mix-APR-P}).
\textit{AF-APR-P} swaps the phase spectrum of images as follows:
\begin{equation}\small
	\setlength{\abovedisplayskip}{3pt}
	\setlength{\belowdisplayskip}{3pt}
	x_{j,aug} = iDFT(\mathcal{A}_{x_j} \otimes e^{i \cdot \mathcal{P}_{x_k}}),
	\label{eqn:freq}
\end{equation}
Therefore, the ground truth label corresponding to the original image's amplitude spectrum is preserved.
\textit{Mix-APR-P} randomly selects the spectral component from which to assign the respective label, forcing the network to extract both phase and amplitude features.

Table~\ref{tab:aprp-results} shows the test performance of models trained with different variations of \textit{APR-P} augmentation.
\textit{AF-APR-P} outperformed the other variations in the case of binary skin lesion classification, although statistical significance is only achieved in case of \textit{ISIC-b}.
When classifying skin lesions in multiple disease classes, neither augmentation showed any benefit.
However, it can be seen that \textit{APR-P} decreased the average test accuracy in all skin classification tasks.
For both visual recognition datasets, neither \textit{APR} augmentation improved the results.
As expected, \textit{AF-APR-P} and \textit{Mix-APR-P} even led to a significant decrease in most cases.

\begin{table}[t!]
\caption{Average test results of random retraining with different variants of \textit{APR} augmentation. Statistical significance of average accuracy to the baseline training is indicated by asterisks.}
\label{tab:aprp-results}
\begin{tabular}{p{0.18\textwidth}p{0.1\textwidth}p{0.12\textwidth}p{0.1\textwidth}p{0.1\textwidth}p{0.17\textwidth}p{0.17\textwidth}}
\toprule
\textbf{Dataset} & \textbf{D7P-b} & \textbf{ISIC-b} & \textbf{D7P} & \textbf{ISIC} & \textbf{Imagenette} & \textbf{Imagewoof} \\ \midrule
Baseline        & 82.01          & 88.45           & \textbf{69.13}        & \textbf{85.93}         & \textbf{97.77}               & \textbf{92.53}              \\ \cline{2-7}
APR-P            & 80.4*           & 88.62           & 66.05*        & 82.79         & 96.52               & 91.47              \\
AF-APR-P         & \textbf{82.42}          & \textbf{89.23**}           & 67.43        & 84.18         & 97.06**               & 91.91***              \\
MIX-APR-P        & 81.5           & 89.16           & 68.62        & 84.22         & 97.3**                & 91.68             \\ \bottomrule
\addlinespace[1ex]
\multicolumn{3}{l}{\textsuperscript{***}$p<0.01$, 
  \textsuperscript{**}$p<0.05$, 
  \textsuperscript{*}$p<0.1$}
\end{tabular}
\end{table}

\section{Discussion}
\label{sec:discussion}
We have shown there exist different, dataset-dependend biases with respect to \textit{Texture}, \textit{Shape} and \textit{Color} features in dermoscopy.
\textit{D7P} datasets seem to be more biased towards \textit{Shape} as compared to \textit{Texture}.
One possible explanation for this finding is the significant difference in training size between \textit{ISIC} and \textit{D7P} ($\times 28$ for binary and $\times 53$ for multiclass) allows the \textit{ISIC}-trained models to pick up more nuanced features including fine-grained textures.
This effect requires further investigation and is of special importance for the robustness and explainability of skin lesion classifiers in clinical use.

The results in section~\ref{sec:dfr} show differences between the feature entanglement of binary and multiclass classification tasks and indicated that a partial or even full encoding of disentangled features is possible depending on the complexity of the target task.
However, it has also been shown that the end-to-end trained classifier does not necessarily use these features independently, potentially suffering a loss in robustness.
Additional experimentation is required to investigate potential mechanisms leading to disentangled classification layers from end-to-end training. 
One possible way would be the explicit data augmentation with the different feature isolations proposed in section~\ref{sec:background}.
Another interesting direction would be the application of contrastive losses from self-supervised learning to achieve a better alignment of different feature-isolated ablations.

Section~\ref{sec:phase-amplitude-dependence} revealed that features relevant for dermoscopic classification are spread across different components in the spectral domain, whereas visual recognition classifiers are deliberately biased towards only the phase component in order to increase robustness.
A reason for this phenomenon might lie in the relevance of \textit{Texture}, \textit{Shape} and \textit{Color} in skin lesion classification.
The phase spectrum is known to encode mainly edges, which often correspond to coarse structures as outlines, but also fine-grained structures resembling textures.
On the other hand, color is mainly encoded in the amplitude spectrum, as it contains information about the magnitude of specific frequency components in the respective color channels.

%APR-P
These findings indicated that common methods for increased robustness, which reinforce the \textit{shape-bias} are not necessarily suitable for skin lesion classification.
However, skin lesion classification has been shown to significantly benefit from our proposed Amplitude-Focused \textit{APR} augmentation.
Multiclass skin lesion classification seem to not benefit, or even suffer from the Fourier-domain augmentation.
A reason for this behaviour might be the increasing relevance of \textit{Texture} and \textit{Shape} features in multiclass settings, as reported in section~\ref{sec:dfr}.
The intuition behind the \textit{Mix-APR-P} augmentation was that a random exposure to phase- or amplitude-randomized samples might implicitly force the network to learn individual features.
Yet, this assumption has been disproved.

Overall, the findings of this work indicate an inherent complexity of the dermoscopic skin lesion classification task.
Indeed, the process of clinical reasoning has already been shown to be fundamentally different from other human decision such as visual recognition, based mostly on analytical reasoning.
Norman et al.~\cite{norman2009iterative} describe the process of clinical reasoning as an iterative approach, combining non-analytical with analytical operations to varying degrees, depending upon personal style preferences, experience, and awareness of the diagnostic task~\cite{dinnes2018visual}.
Commonly applied diagnostic procedures in dermoscopy are manual algorithms like the ABCD rule~\cite{stolz1994abcd}, the Menzies method~\cite{menzies1996frequency}, the seven-point checklist~\cite{argenziano1998epiluminescence} as well as the method of pattern analysis~\cite{kittler2016dermatoscopy}.
Methods like the seven-point checklist and pattern analysis are based on the identification of complex dermoscopic features such as Blue-Whitish-Veil or Pigment Networks.
The successful application of pattern analysis requires years of extensive training.
Evidence suggest, that clinical reasoning in dermoscopy puts more emphasis on the non-analytical, unconscious description of overall patterns as compared to analytical processes~\cite{gachon2005first,zalaudek2008time}.

This suggests that dermoscopic skin lesion classification and other medical imaging tasks pose particularly interesting challenges upon the whole computer vision community.
Due to the inherent complexity of the task, special efforts are required from the community working on explainable Artificial Intelligence, in order to properly disentangle features and align explanations with the human clinical reasoning processes.

\section{Conclusion}
\label{sec:conclusion}
In this paper, we revisit the utility of developing \textit{shape-biased} models for recognition beyond natural images.
Particularly, we consider the domain of dermoscopic skin lesion classification.
Through a range of different experiments, we have shown that deep features learnt for the classification of skin lesions are inherently entangled due to the complexity of the underlying task.
At the same time, our analysis reveals that feature disentanglement can be achieved even on networks trained without constraints, and found that an increasing task complexity as well as a higher number of training samples leads deep feature extractors to learn a more diverse set of redundant and isolated features.
Additionally, we showed that dermoscopic features are spread over different spectral components in contrast to common visual recognition tasks like \textit{Imagenet}.
This indicates that the commonly desired \textit{shape-bias} for improved model robustness does not apply in dermoscopy, and that the task requires specifically tailored solutions.
We demonstrated a first step towards dermoscopy-specific robustness measures beyond \textit{shape-bias} by introducing Amplitude-Focused Amplitude-Phase Recombination, showing improved performance on binary skin lesion classification tasks.
More importantly, this work highlights the importance of scrutinizing a given computer vision task in order to find relevant, and robust requirements for the decision-making.
Dermoscopy is only one out of plenty use-cases with unique requirements which extend beyond the simple analytical procedures of visual object recognition.
These kind of considerations are particularly important for pivotal areas such as self-supervised learning, where an adequate requirement engineering will potentially lead to enormous performance improvements.

%%%%%%%%%%%%%%
% References %
%%%%%%%%%%%%%%

\bibliographystyle{splncs04}
\bibliography{main.bib}

\end{document}